\documentclass[twoside]{article}

\usepackage[accepted]{paper}
\usepackage{graphicx}
\usepackage{xspace}
\usepackage{enumitem}
\usepackage[english]{babel}
\usepackage{amsthm}
\usepackage{booktabs} 
\usepackage{amssymb}
\usepackage{pifont}  
\usepackage{url}
\usepackage{mathrsfs}
\usepackage{float}

\newtheorem{theorem}{Theorem}
\newtheorem{corollary}{Corollary}

\newtheorem{remark}{Remark}

\usepackage[round]{natbib}

\newcommand{\methodname}{\mbox{SPICE}\@\xspace}
\newcommand{\methodnameHPD}{\mbox{SPICE-HPD}\@\xspace}
\newcommand{\methodnameNegDensity}{\mbox{SPICE-ND}\@\xspace}
\newcommand{\capsmethodname}{\mbox{SPICE}\@\xspace}
\newcommand{\encoder}{\textsc{NN$_{\theta_e}$}}

\newcommand{\knotHeightLinear}{\ensuremath{\{h_i\}_{i=1}^K}}

\newcommand{\knotpos}{\ensuremath{\{t_i\}_{i=1}^K}}
\newcommand{\predSet}{\ensuremath{\mathcal{C}}}
\newcommand{\predSetFhat}{\ensuremath{\mathcal{C}_\textsc{ND}}}
\newcommand{\predSetHPD}{\ensuremath{\mathcal{C}_\textsc{HPD}}}
\newcommand{\estimDensity}{\ensuremath{\hat f_\theta}}
\newcommand{\score}{\mathcal{S}}
\newcommand{\scoreFhat}{\ensuremath{\mathcal{S}_{\hat f}}}
\newcommand{\scoreHPD}{\ensuremath{\mathcal{S}_{\textsc{HPD}}}}
\newcommand{\trainData}{\ensuremath{\mathcal{D}_\text{train}}}
\newcommand{\calData}{\ensuremath{\mathcal{D}_\text{cal}}}

\newcommand{\unitInt}{\ensuremath{\mathbb{I}}}

\begin{document}

\runningtitle{Conformalized Deep Splines}

\twocolumn[

\papertitle{Conformalized Deep Splines for \\ Optimal and Efficient Prediction Sets}

\paperauthor{ Nathaniel Diamant \And Ehsan Hajiramezanali \And Tommaso Biancalani \And Gabriele Scalia }

\paperaddress{ Biology Research $|$ AI Development (BRAID), Genentech } ]

\begin{abstract}
    Uncertainty estimation is critical in high-stakes machine learning applications. 
    One effective way to estimate uncertainty is conformal prediction, which can provide predictive inference with statistical coverage guarantees. 
    We present a new conformal regression method, \emph{Spline Prediction Intervals via Conformal Estimation} (SPICE), that estimates the conditional density using neural-network-parameterized splines.
    We prove universal approximation and optimality results for SPICE, which are empirically validated by our experiments.
    \methodname is compatible with two different efficient-to-compute conformal scores, one oracle-optimal for marginal coverage (\methodnameNegDensity) and the other asymptotically optimal for conditional coverage (\methodnameHPD).
    Results on benchmark datasets demonstrate \methodnameNegDensity models achieve the smallest average prediction set sizes, including average size reductions of nearly 50\% for some datasets compared to the next best baseline.
    \methodnameHPD models achieve the best conditional coverage compared to baselines.
    The SPICE implementation is made available.
\end{abstract}

\begin{table*}[!t]
\caption{Comparison of the properties of different conformal regression methods. \textsc{disjoint} indicates whether a method can output multiple prediction intervals. \textsc{opt. marginal} indicates whether a method has oracle-optimal marginal size (see Section~\ref{sec:desiderata} for more information). \textsc{opt. conditional} indicates asymptotically optimal conditional size. \textsc{deterministic} indicates whether prediction set computation is deterministic or stochastic. \textsc{computation} is the computational complexity of computing a prediction set. $^*$CHR has asymptotically optimal conditional interval size only for unimodal conditional distributions.
} \label{tab:desiderata}
\begin{center}
\resizebox{\textwidth}{!}{
    \begin{tabular}{lcccccc}
    \textsc{MODEL} & \textsc{disjoint} & \textsc{opt. marginal} & \textsc{opt. conditional} & \textsc{deterministic} & \textsc{computation} \\
    \midrule
    \textsc{CQR} \citep{romano_conformalized_2019} & \ding{56} & \ding{56} &  \ding{56} &  \ding{52} & $\mathcal{O}(1)$ \\
    \textsc{CHR} \citep{sesia_conformal_2021} & \ding{56} & \ding{56} &  \hspace{4pt}\ding{52}$^*$& \ding{56}  & $\mathcal{O}([\text{\# bins $\approx 10^3$}] \cdot [\text{\# grid points = 1,000}])$ \\
    \textsc{PCP} \citep{wang_probabilistic_2023} & \ding{52} & {\large \textbf{?}} &  {\large \textbf{?}} & \ding{56} & $\mathcal{O}([\text{\# samples} \approx 50] \cdot [\text{sample complexity}])$ \\
    \textsc{\methodnameNegDensity [OURS]} & \ding{52} & \ding{52} &  \ding{56} & \ding{52} & $\mathcal{O}([\text{\# knots} \approx 30])$\\
    \textsc{\methodnameHPD [OURS]} & \ding{52} & \ding{56} &  \ding{52} & \ding{52}  & $\mathcal{O}(\text{\# knots} \cdot [\log \text{precision} \approx 15])$ \\
    \end{tabular}
}
\end{center}
\end{table*}

\section{INTRODUCTION}

Uncertainty estimation is an essential problem in machine learning, especially for domains with high-stakes decisions such as medicine \citep{begoli_need_2019}, drug discovery \citep{scalia_evaluating_2020}, and self-driving \citep{abdar_review_2021}.
This work focuses on predictive inference for scalar regression, where the goal is to build a \emph{predictive set} rather than output a single value, in which the true target value is likely to fall. This formulation is an effective way to account for predictive uncertainty.

Conformal prediction~\citep{vovk2005algorithmic} is one of the most successful frameworks to approach building predictive sets. It only requires exchangeability of the data, does not make distributional and model assumptions, and comes with non-asymptotic coverage guarantee.

In this work, we present a deep regression model specifically designed for conformal prediction techniques, resulting in small and efficient-to-compute prediction sets with statistical coverage guarantees. As we will show, our proposed method has several theoretical and practical advantages over existing conformal prediction techniques and results in sharper predictive sets and improved conditional coverage on a range of benchmark datasets. 

\subsection{Conformal Prediction Background}
We present an overview of split conformal prediction \citep{lei_conformal_2015} for completeness and to introduce notation.
For a more comprehensive introduction, see \citet{angelopoulos_gentle_2022}.
Let $\{(x_i, y_i)\}_{i=1}^N$ denote $N$ samples drawn from an unknown joint distribution with density function $f_{X, Y}(x, y)$.
In split conformal, we randomly split the data into a training set $\trainData = \{(x_i, y_i)\}_{i=1}^{N_\text{train}}$ and calibration data $\calData = \{(x_i, y_i)\}_{i=N_\text{train} + 1}^{N_\text{train} + N_\text{cal}}$.

\subsubsection{Conformal Scores}\label{sec:conformal_score_intro}
The first step is to train a model $\hat f(x)$ on $\trainData$.
Given the model, we define a conformal score $\score(x, y)$, which is higher the more the model's prediction disagrees with $y$.
In a regression setting, a common way to build a conformal score is to divide the absolute residual by some heuristic uncertainty score, $u(x)$, such as the ensemble variance \citep[\S2.3]{angelopoulos_gentle_2022}.
In the case where $\hat f(x)$ models the conditional density of $Y \mid X$, one can use $\score(x, y) = -\hat f(y \mid x)$.
No matter what model and conformal score, the conformal prediction sets will have statistical guarantees.
However, the modeling and score choices are critical for the size and properties of the prediction sets.
A poor choice of model or score will lead to large and uninformative prediction sets.

\subsubsection{Conformal Prediction Sets}\label{sec:pred_sets_intro}

In conformal prediction, the user specifies a desired miscoverage rate, $\alpha \in [0, 1]$.
$1 - \alpha$ is called the \textit{nominal coverage}.
Assuming $(x_\text{test}, y_\text{test})$ and $\calData$ are exchangeable, split conformal builds prediction sets $\predSet(x)$ such that:
\begin{equation}\label{eq:marginal_cov}
    \mathbb{P}[y_\text{test} \in \predSet(x_\text{test})] \geq 1 - \alpha,
\end{equation}
where the probability is taken over all possible draws of $(x_\text{test}, y_\text{test})$ and $\calData$.
This property is known as \textit{marginal coverage}, since for any individual $(x_\text{test}, y_\text{test})$ pair we have no guarantee.
The stronger \textit{conditional coverage} would be if Equation~\ref{eq:marginal_cov} held for each specific test sample.

In order to build the prediction sets, we calculate a quantile of the conformal scores of the calibration set, which we denote $\score_\text{cal} =  \{\score(x, y) \colon (x, y) \in \calData \}$, as follows:
\begin{equation}\label{eq:calibration-quantile}
    \hat q = \textsc{quantile} \left[\score_\text{cal}, \frac{\lceil (N_\text{cal} + 1)(1 - \alpha) \rceil)}{N_\text{cal}} \right].
\end{equation}
Equation~\ref{eq:calibration-quantile} is a slightly adjusted version of the nominal coverage quantile of the conformal score.
The prediction set for a test sample is then:
\begin{equation}\label{eq:pred_set}
    \predSet(x_\text{test}) = \{y \colon \score(x_\text{test}, y) \leq \hat q\}.
\end{equation}
Intuitively, the prediction set contains all the possible values of $y$ that align well, to a degree determined by $\hat q$, with the model's prediction.
A more accurate model will have a smaller $\hat q$, reflecting the fact that the true $y$s better align with the model's predictions in the calibration set.

\subsection{Conformal Regression Desiderata}\label{sec:desiderata}
We have seen that any conformal score will result in marginal coverage. Therefore, we have to compare conformal methods based on other important properties of the prediction sets.

One key property is the \emph{flexibility} of $\predSet(x)$ in its ability to cover multiple intervals of possible values for $y$.
\citet{sesia_conformal_2021} argue that the prediction set should be a single interval for improved interpretability.
While this may be beneficial in some contexts, outputting multiple prediction intervals can make the prediction set much smaller in many cases \citep{wang_probabilistic_2023}.
One case where this may be important is if there is a hidden categorical variable that leads to a multimodal conditional distribution for $y$, in which case a single interval could have to span between multiple modes and thereby include low-density regions. Additionally, applications with inherently multimodal observations have been highlighted across many domains~\citep{izbicki_cd-split_2022}.

Another important design principle is the size of the prediction intervals, assuming a perfect conditional density model (i.e. $\hat f(y \mid x) = f(y \mid x)$).
Intuitively, given an oracle model, a well-designed conformal score should result in the smallest possible prediction intervals \citep{lei2014distribution}.
We refer to this property, without any constraints besides marginal coverage, as \textit{optimal marginal size}.
When the score also results in conditional coverage, we call the property \textit{optimal conditional size}.
Optimal marginal size is achieved by under-covering highly variable conditional examples, which means that both properties cannot be simultaneously achieved \citep{sadinle2019least}.
In practice, optimal conditional coverage can lead to improved non-oracle conditional coverage, while resulting in larger average prediction set sizes.
One example of a regression score with optimal conditional coverage is the High Predictive Density (HPD) score \citep{izbicki_cd-split_2022}.

\begin{figure*}[t]
\includegraphics[width=\linewidth]{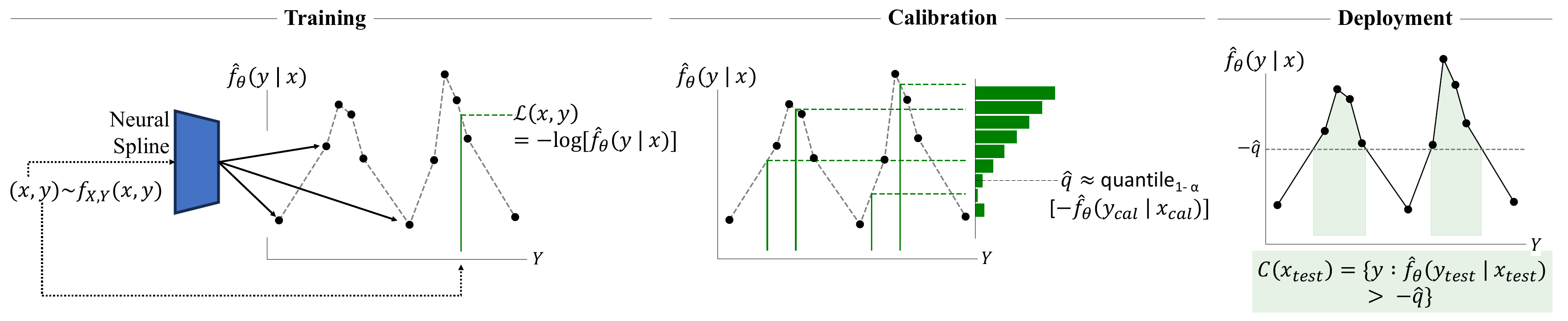}
\caption{The three stages of \methodnameNegDensity. \methodnameHPD shares the first stage but differs in the calibration and prediction stages.}
\label{fig:overview}
\end{figure*}

Another practical consideration is the \emph{complexity} of computing the conformal score and prediction sets.
Some models, such as Conformalized Quantile Regression (CQR) \citep{romano_conformalized_2019}, only require adding an offset to a prediction interval.
Others require numeric integration \citep{izbicki_cd-split_2022}, random sampling \citep{wang_probabilistic_2023}, or constructing large nested sets of intervals \citep{sesia_conformal_2021}. A high computational complexity can hinder applications in real-world settings. Additionally, some methods use randomized procedures to build the predictive set, resulting in stochastic predictions. In this regard, a \emph{deterministic} computation could improve reproducibility, which is critical in many domains.

We summarize these desiderata in Table~\ref{tab:desiderata} and show that our method, \emph{Spline Prediction Intervals via Conformal Estimation} (SPICE), achieves a uniquely strong combination of them.

Overall, we highlight the following contributions:
\begin{itemize}[itemsep=0em]
    \item We introduce a new and effective conformal regression method (\methodname), with two conformal score variants that can be efficiently computed on GPU, and release the code\footnote{\url{https://github.com/ndiamant/spice}}.
    \item \methodname prediction sets with the negative conditional density conformal score (\methodnameNegDensity) have oracle-optimal marginal size.
    \item \methodname prediction sets with the HPD conformal score (\methodnameHPD) have asymptotically optimal conditional size.
    \item We prove that \methodname can uniformly approximate any continuous conditional density function and that it can express any finite union of prediction intervals (Section~\ref{sec:universal_slpine}, \ref{sec:pred_sets_thm}).
    \item \methodnameNegDensity models consistently achieve the smallest prediction sets compared to all baselines on standard conformal regression benchmarks.
    \item \methodnameHPD models achieve the best conditional coverage results compared to the baseline models.
\end{itemize}

\section{Spline Prediction Intervals via
Conformal Estimation (SPICE)}\label{sec:methods}

\subsection{Overview}

\capsmethodname uses a neural network to build a spline estimate of the conditional density $f(Y \mid X)$.
The conditional density estimate can then be used in either of two conformal scores to build prediction sets with different properties.
Overall, \capsmethodname consists of three main components:
\begin{enumerate}
    \item A flexible neural network encoder which converts inputs into embeddings. This is standard across neural conformal-regression methods and enables regression on any modality a neural network can take as input.
    \item A spline module which converts embeddings into conditional density functions (Section~\ref{sec:conditional_density}). This is the central innovation of \methodname. Low-order-splines as density functions enable efficient and closed-form computation of conformal scores prediction sets, while maintaining expressiveness.
    \item One of two conformal scores (Section~\ref{sec:conformal}). One score results in optimal marginal size (Section~\ref{sec:neg_density}), while the other satisfies optimal conditional size (Section~\ref{sec:hpd_score}). Both allow for expressive and flexible prediction sets.
\end{enumerate}
See Figure~\ref{fig:overview} for a visual overview of \methodname.

\subsection{Conditional Density Estimation}\label{sec:conditional_density}
\capsmethodname estimates the conditional density as fourth-degree-and-below polynomial splines, enabling integration and root-finding in closed form.
The degree of a spline is the order of its polynomials, e.g. degree $n = 2$ for quadratic polynomials.
For this work, we used first and second-degree splines, although the same approach can be applied to third and fourth-order polynomials, as they also have closed-form solutions.
We make the assumption that $Y$ has support only on a finite interval, and in practice we zero-one normalize the data so that $0 \leq y < 1$.
For generality, \capsmethodname differentiably maps covariates $x$ to a spline in two stages.
First, an encoder neural network $\encoder(x) = z$ encodes $x$.
Next, two neural network modules map $z$ to $K$ sorted spline-knot positions and $n (K - 1) + 1$ knot heights, where $n$ is the polynomial order.
The differentiable parameterization of positions and heights is detailed in Appendix~\ref{app:knot_encoding}.

\begin{figure}[ht]
\includegraphics[width=\linewidth]{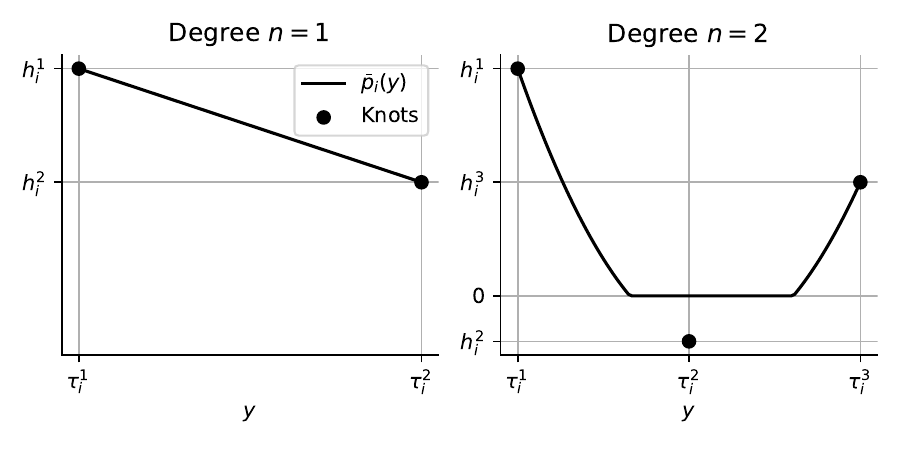}
\caption{Zero-truncated Lagrange polynomials interpolating two (left) and three (right) points.}
\label{fig:interp}
\end{figure}

In this spline parameterization, there is a polynomial between every pair of consecutive knot positions $(t_i, t_{i+1})$.
The positions and heights are used to construct $n + 1$ points to uniquely specify each polynomial.
We index the heights used to specify the $i$-th polynomial as $h_i^j$, where $j$ ranges in $(1, \ldots, n + 1)$. 
Note that some heights are redundantly indexed; i.e. $h_i^{n + 1} = h_{i + 1}^1$.
In addition to the knots, we need $n - 1$ evenly\footnote{For degrees three and four, Chebyshev points would likely be more effective \citep{trefethen2019approximation}.} spaced intermediate positions between $t_i$ and $t_{i + 1}$ (Equation~\ref{eq:intermediate_knots}):
\begin{equation}\label{eq:intermediate_knots}
    \tau_i^j = t_i + (j - 1) \cdot \frac{(t_{i + 1} - t_{i})}{n} \ \text{, where $j = 1, \ldots, n + 1$}.
\end{equation}
Using the positions and heights, we derive the unique order-$n$ Lagrange polynomial, $p_i(y)$, that interpolates $\{(\tau_i^j, h_i^j)\}_{j=1}^{n + 1}$.
We constrain $h_i^1$ and $h_i^{n + 1}$ to be positive, but let all other $h_i^j$ be any real number to maximize expressiveness.
In order to ensure $p_i(y) > 0$, a requirement to build a density function, we define $\bar p_i(y) = \max\left(p_i(y), 0\right)$.
Figure~\ref{fig:interp} illustrates examples for degrees one and two.

\subsubsection{Normalizing the Conditional Density}\label{sec:normalization}
So far, we have shown how \methodname builds a non-negative spline from $x$.
In order for the spline to be a valid density function, it needs to be normalized such that its integral evaluates to one.
The normalizing constant is simply the sum of the integrals of the zero-truncated polynomials.
The normalizing constant can be efficiently computed on GPU in $\mathcal{O}(Kn)$ time by integrating each polynomial $p_i(y)$ between its bounds and then subtracting the integrals between the polynomials' roots (Appendix~\ref{app:integration}).
The final normalized conditional density estimate is:
\begin{equation}\label{eq:estim_conditional_density}
    \estimDensity(y \mid x) = \bar p_\ell(y)\ /\ \sum_{i=1}^{K - 1}\int_{t_i}^{t_{i + 1}} \bar p_i(y')dy',
\end{equation}
where $\ell$ is selected such that $t_\ell \leq y < t_{\ell + 1}$.
Note that $\ell$ can be found in $\mathcal{O}(\log K)$ time using binary search, since the knot positions are sorted, resulting in an overall complexity in $\mathcal{O}(Kn)$ for evaluating the estimated conditional density.

\subsubsection{Training the Conditional Density Estimator}
$\estimDensity(y \mid x)$ is differentiably parameterized with respect to the encoder, knot position, and knot height parameters ($\{\theta_e, \theta_t, \theta_h\}$).\
Therefore, \capsmethodname can be trained using standard batch gradient descent methods to maximize the log-likelihood of the observed $(x, y) \in \trainData$.

\subsubsection{Universal Approximation Theorem for Neural Splines}\label{sec:universal_slpine}
Both splines and neural networks are known to be very expressive, so it is not surprising that combining them results in a flexible function class.
Without loss of generality to other intervals, we present the following universal approximation result on the unit cube.

\begin{theorem}\label{thm:approximation}
    Let $\unitInt$ denote the unit interval $[0, 1]$.
    Suppose for integer $d > 1$ that $f \colon (\unitInt^{d-1} \times \unitInt) \to \mathbb{R} \geq 0$ is continuous.
    Then for any $\epsilon > 0$, there exists a \methodname model without the unit-integral-normalization step, which we denote $\hat f(x, y) \colon (\unitInt^{d-1} \times \unitInt) \to \mathbb{R} \geq 0$, for which $\sup_{(x, y) \in \unitInt^d} |f(x, y) - \hat f(x, y)| < \epsilon$.
\end{theorem}
Proof of Theorem~\ref{thm:approximation} is provided in Appendix~\ref{app:approximation_thm_proof}.

\begin{remark}
    Theorem~\ref{thm:approximation} may be extensible to non-continuous functions using recent results for approximating discontinuous functions \citep{ISMAILOV2023127096}.
\end{remark}

\begin{corollary}\label{corr:conditional_density_approx}
    Unit-integral-normalized \capsmethodname can approximate any continuous conditional density function $f$ on $[0, 1]^d$.
    This follows from the fact that un-normalized \methodname can approximate $f$ and that the normalizing constant can be made arbitrarily close to one.
\end{corollary}

\subsection{Conformalizing \capsmethodname}\label{sec:conformal}
\capsmethodname builds prediction intervals using either of two conformal scores: $\scoreFhat(x, y) = -\estimDensity(y \mid x)$ (Section~\ref{sec:neg_density}) or the Highest Predictive Density (HPD) score, $\scoreHPD(x, y)$, defined in \citet{izbicki_cd-split_2022} (Section~\ref{sec:hpd_score}).
We refer to \methodname with the negative density score as \methodnameNegDensity, and \methodname with the HPD-score as \methodnameHPD.

\subsubsection{Negative Density Conformal Score}\label{sec:neg_density}
\methodname can use the negative estimated density of $y$, i.e. $\score_{\hat f}(x, y) = -\estimDensity(y \mid x)$, as an efficient and effective conformal score.
As noted before, \methodname can compute the conditional density at $y$ in $\mathcal{O}(Kn)$ time, which is much faster than the forward pass of the neural network encoder for practical values of $K$.
A cutoff $\hat q$ with which to build prediction sets is found using the conformal scores computed on the calibration set as described in Section~\ref{sec:conformal_score_intro}.

\begin{figure}[ht]
\includegraphics[width=\linewidth]{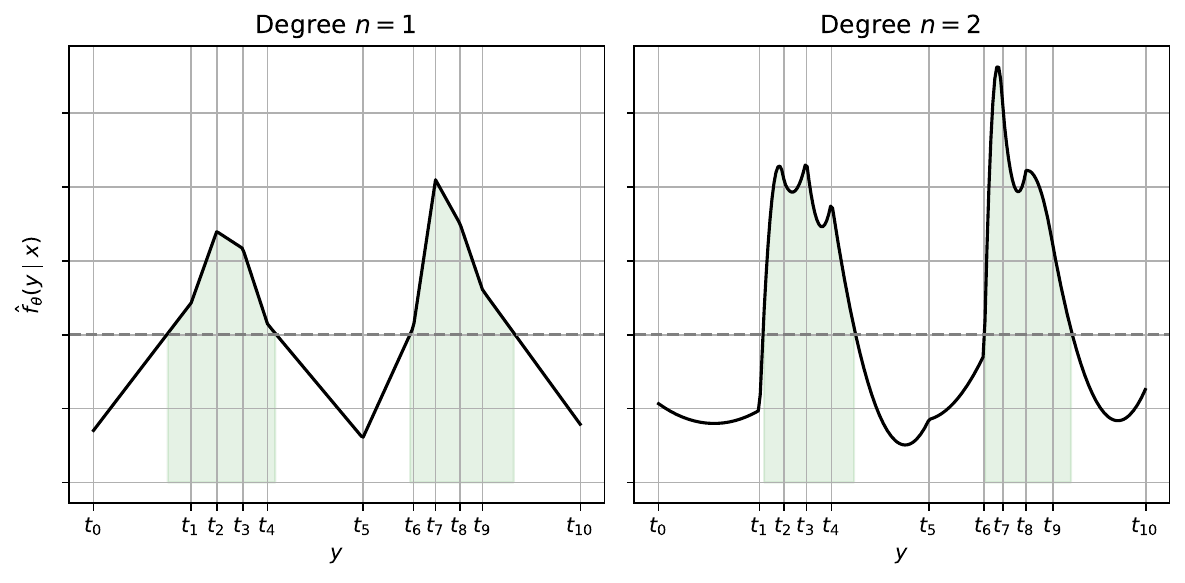}
\caption{Examples of $\predSetFhat(x)$ for first (left) and second degree (right) \methodnameNegDensity. The horizontal dashed line is $-\hat q$, and the filled-in regions show the intervals of $y$ in $\predSetFhat(x)$.}
\label{fig:example_pred}
\end{figure}

Given $\hat q$, the prediction set $\predSetFhat(x)$ consists of all the values $y$ such that $\estimDensity(y \mid x) > -\hat q$, following the conformal procedure described in Section~\ref{sec:pred_sets_intro}.
Figure~\ref{fig:example_pred} shows prediction sets for first and second-degree \methodnameNegDensity.
The prediction set $\predSetFhat(x)$ can be computed in $\mathcal{O}(Kn)$ time since all it requires is finding the roots of low-order polynomials (Appendix~\ref{app:nd_pred_set}).

\methodnameNegDensity achieves oracle-optimal marginal size due to the argument of \citet[\S2.1]{sadinle2019least} as a corollary of the Neyman–Pearson lemma.
Under an argument similar to that of \citet{izbicki_cd-split_2022}, this could likely be extended into asymptotic optimality under consistency assumptions.
However, we believe oracle-optimality is sufficient to justify the design choice of the negative density score.
Indeed, we empirically found the negative density score yielded the smallest average prediction set sizes (Section~\ref{sec:benchmark}).
Interestingly, in a mixed frequentist-Bayesian setting and under some technical assumptions, using the negative-posterior-predictive-density of $y \mid x$ as the conformal score minimizes the Bayes-risk while preserving frequentist marginal coverage \citep{hoff2023bayes}.

\subsubsection{HPD Conformal Score}\label{sec:hpd_score}
Under the assumptions outlined for \citet[Theorem 25]{izbicki_cd-split_2022}, which include the consistency of the conditional density estimator, the HPD-score results in asymptotically optimal conditional prediction set sizes, which is a stronger property than oracle-optimal conditional coverage.
The HPD-score is defined as the negative\footnote{For notational consistency we swapped the sign of the HPD-score from its original presentation.} integral of the estimated conditional density where the density is less than the density of $y$:
\begin{equation}
    \scoreHPD(x, y) = -\int_{y' \colon \estimDensity(y' \mid x) \leq \estimDensity(y \mid x)} \estimDensity(y' \mid x) dy'.
\end{equation}
The HPD-score indicates a higher model error when $y$ has lower density compared to other potential values of $y$.
Notably, \methodname can compute the HPD-score in $\mathcal{O}(Kn)$ using the efficient root finding and integration properties of low-order polynomials.

In order to build prediction sets, a cutoff $\hat q$ is found using the split-conformal procedure described in Sections~\ref{sec:conformal_score_intro} resulting in the following prediction sets:
\begin{equation}
    \predSetHPD(x) = \left\{y \colon \int_{y' \colon \estimDensity(y' \mid x) \leq \estimDensity(y \mid x)} \estimDensity(y' \mid x)dy' > -\hat q \right\}.
\end{equation}
Computing an HPD prediction set for \methodname is more involved than for the negative density score, and is done approximately to arbitrary precision using the bisection method (Appendix~\ref{app:hpd_pred_set}).
This procedure has an asymptotic runtime in $\mathcal{O}\left(Kn \log_2 \frac{1}{\epsilon}\right)$, where $\epsilon$ is the desired precision.
In practice, we only need to run the bisection method for 24 steps to have higher precision than 32-bit floats, and all the computations can be done on GPU.

\subsubsection{Prediction Sets Expressible by \capsmethodname}\label{sec:pred_sets_thm}
Here, we examine which prediction sets \methodname is capable of expressing.
It is clear that given control of the conformal cutoff $\hat q$ and enough knots, \methodname could express any finite union of intervals in $[0, 1]$.
However, unique to \methodname is its flexibility to build any prediction set even without control of $\hat q$.
For example, PCP~\citep{wang_probabilistic_2023} can express up to as many intervals as samples are taken from its conditional generative model (Section~\ref{sec:related_work}), but the smallest any of those intervals can be is $2\hat q$.
That extra flexibility may partially explain the empirical reduction in interval sizes achieved by \methodname (Section~\ref{sec:benchmark}).
We formalize this flexibility in the following theorem.

\begin{theorem}\label{thm:intervals}
    Let $\mathscr{P} = \bigcup_{i=1}^m [a_i, b_i] \subseteq [0, 1]$ where $a_i < b_i < a_{i + 1}$ and $1 \leq m \leq M$ for some natural numbers $m$ and $M$.
    Then the following are true with $K = 4M + 2$:
    \begin{enumerate}
        \item Given any $\hat q \colon 0 < -\hat q \cdot |\mathscr{P}| < 1$ there exists a \methodnameNegDensity model with $K$ knots such that $\predSetFhat(x) = \mathscr{P}$ for some arbitrary $x$.
        \item Given any $0 < -\hat q < 1$ there exists a \methodnameHPD model with $K$ knots such that $\predSetHPD(x) = \mathscr{P}$ for some arbitrary $x$.
    \end{enumerate}
\end{theorem}

See Appendix~\ref{app:intervals_proof} for a proof.
Theorem~\ref{thm:intervals} shows that \methodname can express any union of intervals with a number of knots linear in the number of intervals given any conformal cutoff.

\section{RELATED WORK}\label{sec:related_work}

\citet{durkan_cubic-spline_2019} introduced a monotonically increasing cubic spline, which served as an invertible layer in the context of normalizing flows.
This was part of the inspiration for the spline conditional density function, and we leveraged their technique for differentiably parameterized spline bin-widths.

Conformalized Quantile Regression (CQR) \citep{romano_conformalized_2019} has become one of the most common conformal regression methods.
CQR uses the pinball-loss \citep{steinwart_estimating_2011} to train a neural network to estimate the $\alpha$ and $1 - \alpha$ quantiles of $Y \mid x$, where $\alpha$ is a tunable hyperparameter.
The conformal procedure results in an offset $\hat q$ which leads to prediction set $\predSet(x) = [\hat f_{\alpha}(x) - \hat q, \hat f_{1 - \alpha}(x) + \hat q]$.
Asymptotically, CQR does have conditional coverage, but does not always find the minimal prediction interval \citep{sesia_conformal_2021}.

Conditional Histogram Regression (CHR) \citep{sesia_conformal_2021} can be seen as a more flexible extension of CQR with the additional goal of finding the single minimal prediction interval with asymptotic conditional coverage.
Rather than estimating just two quantiles, CHR estimates many (up to 1,000 in their experiments).
The CHR conformal score involves building a series of nested intervals, each being a subset or superset of an initial interval following the nested sets paradigm \citep{gupta2022nested}.
Each interval is designed to be the minimal interval that is predicted to contain a target probability mass, and that satisfies nesting with the initial interval.
The conformal score is the index of the nested interval that contains the true $y$.
Assuming that the conditional distribution is unimodal and that the quantile prediction model is consistent, CHR asymptotically finds the smallest possible interval and achieves conditional coverage.

Probabilistic Conformal Regression (PCP) \citep{wang_probabilistic_2023} is a method to adapt any conditional generative model to conformal regression.
The PCP conformal score is calculated by sampling from the estimated conditional distribution (usually around 50 samples), and then finding the minimum Euclidean distance between $y$ and any of the samples.
The prediction set is then defined as the union of balls around samples from the target distribution, with the ball radius equal to the quantile computed from the calibration set.
PCP thus enables disjoint prediction intervals, and therefore tends to find smaller prediction sets than CHR or CQR. However, PCP does not provide guarantees for asymptotic conditional coverage. Additionally, both conformal scores and predicted intervals rely on sampled points, making the estimation stochastic.
In the following, we use the strongest PCP variant shown by \citet{wang_probabilistic_2023}, HPD-PCP-MixD, which uses a Gaussian mixture for the conditional distribution and rejects the bottom 20\% least likely samples from the score and prediction set calculations.

Distributional Conformal Prediction (DCP) introduced by \citet{chernozhukov_distributional_2021} is a conformal regression method which uses an estimated conditional cumulative distribution and has similar asymptotic results to CHR.
In experiments, \citet{sesia_conformal_2021} found CHR outperformed DCP, so we did not include DCP as a baseline.
Dist-split and CD-split \citep{izbicki_flexible_2020} are two conformal regression methods that work with conditional density estimators.
These methods were built on by \citet{izbicki_cd-split_2022} to design the HPD conformal score, which we leverage with \methodnameHPD to gain optimal conditional size.
The HPD-score \citep{izbicki_cd-split_2022} was tested with the FlexCode conditional density estimator \citep{izbicki_converting_2017}, which requires many hyperparameters to tune, and was not designed for neural network density estimation, so we did not include it as a baseline.
Both Dist-split and CD-split were outperformed by PCP in \citet{wang_probabilistic_2023}, so we also did not include them as baselines and used PCP as a stronger baseline.

\begin{table*}[ht]
\caption{Normalized marginal size results on benchmark datasets. The nominal coverage was set to 90\%. \textbf{Bold} indicates the best mean; \underline{underline} indicates the second best. \textsc{mean} is the average size across all datasets.} \label{tab:size}
\begin{center}
\begin{small}
\resizebox{\textwidth}{!}{
\renewcommand{\arraystretch}{1.1}
\begin{tabular}{lcccccccc|c}
\textsc{MODEL} & \textsc{bike} & \textsc{bio} & \textsc{blog} & \textsc{meps19} & \textsc{meps20} & \textsc{meps21} & \textsc{star} & \textsc{temp.} & \textsc{mean} \\
\midrule
CQR & $1.00 \pm 0.01$ & $0.99 \pm 0.00$ & $0.20 \pm 0.01$ & $0.98 \pm 0.02$ & $0.91 \pm 0.02$ & $0.87 \pm 0.02$ & $0.89 \pm 0.01$ & $0.89 \pm 0.01$ & $0.84 \pm 0.09$ \\
CHR & $0.89 \pm 0.01$ & $0.98 \pm 0.00$ & $0.14 \pm 0.01$ & $0.74 \pm 0.01$ & $0.70 \pm 0.01$ & $0.66 \pm 0.01$ & $0.91 \pm 0.01$ & $0.89 \pm 0.01$ & $0.74 \pm 0.09$ \\
PCP & $0.23 \pm 0.00$ & $0.59 \pm 0.00$ & $0.16 \pm 0.01$ & $0.81 \pm 0.01$ & $0.82 \pm 0.02$ & $0.79 \pm 0.02$ & $0.89 \pm 0.01$ & $\underline{0.28 \pm 0.00}$ & $0.57 \pm 0.11$ \\
Hist. & $0.26 \pm 0.00$ & $\mathbf{0.53 \pm 0.00}$ & $0.38 \pm 0.00$ & $0.66 \pm 0.03$ & $0.59 \pm 0.07$ & $0.77 \pm 0.01$ & $0.94 \pm 0.01$ & $0.31 \pm 0.01$ & $0.56 \pm 0.08$ \\
\methodnameNegDensity $(n=1$) & $\mathbf{0.20 \pm 0.00}$ & $0.59 \pm 0.02$ & $\mathbf{0.09 \pm 0.00}$ & $0.57 \pm 0.01$ & $\underline{0.55 \pm 0.01}$ & $\underline{0.52 \pm 0.01}$ & $0.90 \pm 0.01$ & $\mathbf{0.27 \pm 0.00}$ & $\underline{0.46 \pm 0.09}$ \\
\methodnameHPD $(n=1$) & $0.24 \pm 0.00$ & $0.62 \pm 0.02$ & $0.15 \pm 0.01$ & $0.87 \pm 0.01$ & $0.82 \pm 0.01$ & $0.87 \pm 0.02$ & $0.93 \pm 0.01$ & $0.29 \pm 0.00$ & $0.60 \pm 0.11$ \\
\methodnameNegDensity $(n=2$) & $\underline{0.22 \pm 0.00}$ & $\underline{0.55 \pm 0.01}$ & $\underline{0.10 \pm 0.00}$ & $\mathbf{0.32 \pm 0.04}$ & $\mathbf{0.35 \pm 0.03}$ & $\mathbf{0.37 \pm 0.03}$ & $\mathbf{0.83 \pm 0.01}$ & $0.28 \pm 0.00$ & $\mathbf{0.38 \pm 0.08}$ \\
\methodnameHPD $(n=2$) & $0.23 \pm 0.00$ & $0.58 \pm 0.01$ & $0.19 \pm 0.01$ & $\underline{0.45 \pm 0.04}$ & $1.11 \pm 0.20$ & $0.64 \pm 0.14$ & $\underline{0.84 \pm 0.01}$ & $0.28 \pm 0.01$ & $0.54 \pm 0.11$ \\
\end{tabular}
}
\end{small}
\end{center}
\end{table*}

\begin{table*}[ht]
\caption{Approximate label-conditional coverage results on benchmark datasets at 90\% nominal coverage. \textbf{Bold} indicates the best mean; \underline{underline} indicates the second best. \textsc{mean} is the average across all datasets.} \label{tab:y_cov}
\begin{center}
\begin{small}
\resizebox{\textwidth}{!}{
\renewcommand{\arraystretch}{1.1}
\begin{tabular}{lcccccccc|c}
\textsc{MODEL} & \textsc{bike} & \textsc{bio} & \textsc{blog} & \textsc{meps19} & \textsc{meps20} & \textsc{meps21} & \textsc{star} & \textsc{temp.} & \textsc{mean} \\
\midrule
CQR & $0.47 \pm 0.01$ & $0.52 \pm 0.00$ & $0.06 \pm 0.01$ & $0.52 \pm 0.00$ & $0.43 \pm 0.01$ & $0.41 \pm 0.01$ & $0.76 \pm 0.01$ & $0.69 \pm 0.01$ & $0.48 \pm 0.07$ \\
CHR & $0.47 \pm 0.01$ & $0.53 \pm 0.00$ & $0.08 \pm 0.01$ & $0.53 \pm 0.00$ & $0.46 \pm 0.01$ & $0.43 \pm 0.01$ & $0.73 \pm 0.01$ & $0.67 \pm 0.01$ & $0.49 \pm 0.07$ \\
PCP & $0.80 \pm 0.01$ & $0.84 \pm 0.00$ & $\underline{0.37 \pm 0.01}$ & $0.53 \pm 0.01$ & $0.51 \pm 0.01$ & $0.50 \pm 0.01$ & $0.81 \pm 0.01$ & $0.87 \pm 0.01$ & $0.65 \pm 0.07$ \\
Hist. & $0.75 \pm 0.01$ & $\mathbf{0.85 \pm 0.00}$ & $0.02 \pm 0.00$ & $0.01 \pm 0.01$ & $0.00 \pm 0.00$ & $0.03 \pm 0.01$ & $0.81 \pm 0.03$ & $0.86 \pm 0.00$ & $0.42 \pm 0.15$ \\
\methodnameNegDensity $(n=1$) & $0.83 \pm 0.01$ & $0.83 \pm 0.01$ & $0.24 \pm 0.01$ & $0.52 \pm 0.01$ & $0.50 \pm 0.01$ & $0.47 \pm 0.01$ & $\underline{0.81 \pm 0.01}$ & $\mathbf{0.87 \pm 0.00}$ & $0.63 \pm 0.08$ \\
\methodnameHPD $(n=1$) & $\mathbf{0.86 \pm 0.00}$ & $0.83 \pm 0.01$ & $0.25 \pm 0.01$ & $\mathbf{0.61 \pm 0.00}$ & $\mathbf{0.60 \pm 0.01}$ & $\mathbf{0.61 \pm 0.01}$ & $\mathbf{0.83 \pm 0.01}$ & $\underline{0.87 \pm 0.00}$ & $\mathbf{0.68 \pm 0.07}$ \\
\methodnameNegDensity $(n=2$) & $0.80 \pm 0.01$ & $0.84 \pm 0.01$ & $0.21 \pm 0.01$ & $0.56 \pm 0.01$ & $0.53 \pm 0.01$ & $0.49 \pm 0.01$ & $0.76 \pm 0.01$ & $0.87 \pm 0.00$ & $0.63 \pm 0.08$ \\
\methodnameHPD $(n=2$) & $\underline{0.85 \pm 0.01}$ & $\underline{0.84 \pm 0.01}$ & $\mathbf{0.42 \pm 0.01}$ & $\underline{0.57 \pm 0.02}$ & $\underline{0.59 \pm 0.02}$ & $\underline{0.54 \pm 0.02}$ & $0.77 \pm 0.01$ & $0.86 \pm 0.01$ & $\underline{0.68 \pm 0.06}$ \\
\end{tabular}
}
\end{small}
\end{center}
\end{table*}

\section{EXPERIMENTS}
We evaluated \methodnameNegDensity, \methodnameHPD, and the baseline models on eight datasets commonly used for conformal regression benchmarking (dataset details in Appendix~\ref{app:benchmark_datasets}).
Each model had approximately the same number of parameters and the same encoder architecture to ensure differences in performance were not caused by architecture.
Hyperparameters were selected using grid search, with the same number of hyperparameter combinations searched over for each method.
Further optimization details are in Appendix~\ref{app:hyperparameters}. 
Results were computed on a heldout test dataset using 20 replicates with different random initializations and train-validation splits for each model (data split details in Appendix~\ref{app:data_split}).

\subsection{Discretized $y$ Baseline}\label{sec:discrete_baseline}
We also introduce a simple baseline for conformal regression by discretizing the domain of $y$ into equally sized bins, and using a conformal technique commonly used in the classification setting.
Specifically, we used the negative model-predicted probability of the discretized $y$ as the conformal score, which is the optimal choice in terms of expected prediction set size for an oracle classifier \citep{sadinle2019least}.
To the best of our knowledge, other studies on conformal regression have not incorporated this baseline. Given its simplicity, properties, and adequate performance, we believe that including it should become common practice in future works.
This model is denoted as \emph{Hist.} in the following results.
Further details are in Appendix~\ref{app:discretized_baseline}.

\subsection{Marginal Size Results}\label{sec:benchmark}
All models achieved their marginal coverage targets as expected, so we compared the average interval sizes on heldout data.
We normalized the size measurement to the size achieved by a marginal histogram of the training $y$ in order to make sizes more comparable across datasets (Appendix~\ref{app:size_normalization}).
As shown in Table~\ref{tab:size}, \methodnameNegDensity with $n = 1, 2$ consistently resulted in the smallest interval sizes.
\methodnameHPD performed worse in terms of average interval size, but was competitive with the baseline methods.
Of the baseline methods, PCP and our histogram baseline were the most performant.
Overall, \methodnameNegDensity $(n = 2)$ had the smallest intervals, perhaps due to the increased flexibility of quadratic versus linear interpolation.

\subsection{Conditional Coverage Results}
Conditional coverage is the coverage of a prediction set for a specific set of covariates, $x$.
It is inaccessible except for synthetic data, so it can only be approximated for real data.
Following the recommendation of \citet{angelopoulos_gentle_2022}, we measured \emph{label-conditional coverage} as our conditional coverage metric (details in Appendix~\ref{app:label_cond_coverage}).

As shown in Table~\ref{tab:y_cov}, \methodnameHPD with $n = 1$ and $n = 2$ had the best and second best conditional coverage respectively on average across all datasets.
In six out of eight datasets, \methodnameHPD outperformed \methodnameNegDensity and otherwise had close to the same conditional coverage, reflecting the benefit of the optimal conditional size property.

\begin{figure*}[ht]
\includegraphics[width=\linewidth]{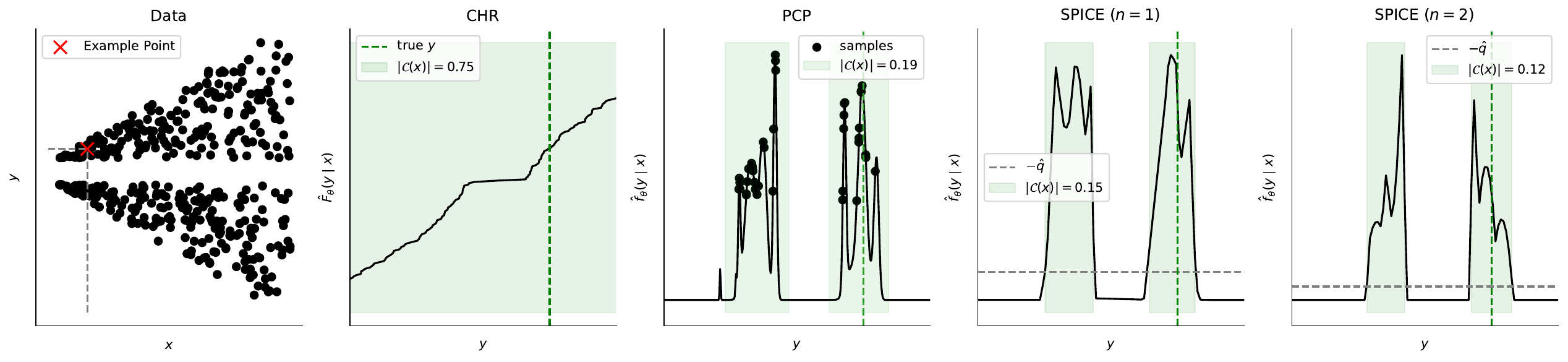}
\caption{Example prediction sets for \methodnameNegDensity and two baselines on a synthetic data sample.}
\label{fig:synth_example}
\end{figure*}

\subsection{Analysis of Interval Sizes}
\begin{figure}[ht]
\includegraphics[width=0.8\linewidth]{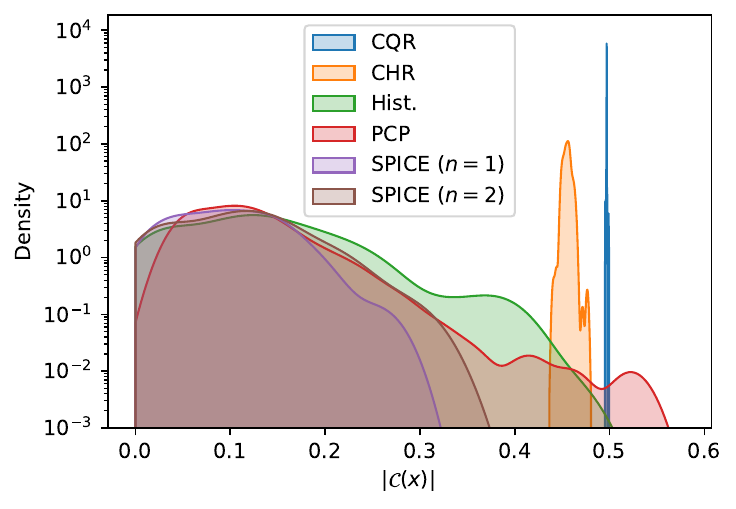}
\caption{Kernel-density plots of the sizes of predictive sets for each model on the \textsc{BIKE} dataset.}
\label{fig:size_dist}
\end{figure}

In order to understand the smaller predictive set sizes of \methodname, we analyzed the distributions of the set sizes for each model on the \textsc{BIKE} dataset (Figure~\ref{fig:size_dist}).
The CQR and CHR set sizes were relatively invariant across test samples compared to the methods with more flexible prediction sets.
PCP had much more varied sizes, potentially reflecting the stochasticity of its predictive set construction procedure.
PCP was also less likely to express the smallest intervals, which may be due to the restriction of the smallest possible interval size it can express (Section~\ref{sec:pred_sets_thm}).

These differences were supported by results on synthetic data with heteroscedastic bimodal conditional distributions (described in Appendix~\ref{app:synthetic_hetero}). We show example prediction sets in Figure~\ref{fig:synth_example}.
We found that CHR and CQR were forced to cover low-density regions due to their single-interval prediction set constraint.
PCP prediction intervals captured the bimodality, but also covered some low predictive density regions.
This is due to the $\hat q$ radius around conditional samples extending outside of the high-density regions.
The \methodnameNegDensity prediction sets covered only high-density regions and thus were consistently smaller than the baselines'.

\section{DISCUSSION}

\methodnameNegDensity consistently achieved the smallest average prediction set sizes across experiments, and \methodnameHPD the best conditional coverage results, reflecting their optimal marginal and conditional size properties.
Nonetheless, \methodnameNegDensity also performed well in terms of conditional coverage. This suggests that \methodnameNegDensity may be a suitable general-purpose choice when marginal coverage is key and conditional coverage is not critical, also because of its computational efficiency (Table~\ref{tab:desiderata}).
Using degree-two splines, \methodname models resulted in smaller interval sizes and comparable conditional coverage results.
The flexibility of quadratic functions compared to linear ones may explain this result, and suggests cubic splines could be a valuable avenue for future exploration.

The strong performance of Hist., the discrete classifier baseline (Section~\ref{sec:discrete_baseline}), highlights the importance of incorporating competitive baselines when evaluating new conformal methods, and should be considered in future conformal regression studies.

A limitation of \methodname is the inability to directly handle multivariate regression problems. An interesting area of future work is extending \methodname to the multivariate case, perhaps by modeling the joint density. For multivariate problems, we currently recommend PCP, despite its lack of optimality results, because of its flexibility of choice in conditional generative models and simple prediction set computation. 

\section{CONCLUSION}

We introduced \methodname, which provides an expressive, efficient, and practical method for uncertainty estimation in scalar regression problems. \methodname achieves a unique combination of advantages and consistently outperforms existing methods, resulting in sharper predictive sets and improved conditional coverage. 
The effectiveness of \methodname highlights the benefits of co-designing the regression model and conformal scores.
The code release will support adoption in applications such as medicine and drug discovery, where reliable uncertainty estimation is critical.
An important direction for future work is applying SPICE to selection problems, such as portfolio optimization and experimental design \citep{jin2023selection,pmlr-v206-stanton23a}. 
Another promising application area is survival analysis \citep{wang2019machine}, where conditional coverage is particularly valuable, and efficient evaluation of the conditional CDF could enable survival predictions with statistical coverage guarantees.

\bibliography{main}
\bibliographystyle{plainnat}


\onecolumn
\appendix
\papertitle{SPICE: Supplementary Materials}

\section{PROOFS}

\subsection{Proof of Universal Approximation Theorem}\label{app:approximation_thm_proof}
We reproduce the theorem here for easy reference.

\begin{theorem}\label{thm:app_universal}
    Let $\unitInt$ denote the unit interval $[0, 1]$.
    Suppose for integer $d > 1$ that $f \colon (\unitInt^{d-1} \times \unitInt) \to \mathbb{R} \geq 0$ is continuous.
    Then for any $\epsilon > 0$, there exists a \methodname model without the normalization to unit-integral step, which we denote $\hat f(x, y) \colon (\unitInt^{d-1} \times \unitInt) \to \mathbb{R} \geq 0$, for which $\sup_{(x, y) \in \unitInt^d} |f(x, y) - \hat f(x, y)| < \epsilon$.
\end{theorem}

This theorem is general to single-hidden-layer neural network encoders for the spline-knot positions and heights.
It also holds for our specific parameterization of spline knots (Appendix~\ref{app:knot_encoding}), since the softmax is a continuous, surjective function onto $(0, 1)$.

\begin{proof}
Let $\phi \in \mathbb{R}^K$ be a vector of spline parameters for natural number $K$ and let $s(\phi, y)$ for $y \in \unitInt$ be the resulting spline function.
Let $\mathscr{S}(x) \colon \unitInt^{d-1} \to \mathbb{R}^K$ be a function mapping $x$ to spline parameters such that for all $x \in \unitInt^{d-1}$
\begin{equation}\label{eq:proof_spline_approx}
    \sup_{y \in \unitInt} |f(x, y) - s(\mathscr{S}(x), y)| < \epsilon / 3.
\end{equation}
We know $\mathscr{S}(x)$ exists for sufficiently large $K$ by the fact that splines can uniformly approximate continuous functions \citep[Theorem 20.11]{powell_1981}.

Next, we define function $E \colon (\unitInt^{d-1} \times \mathbb{R}^K) \to \mathbb{R} \geq 0$ as 
\begin{equation}
    E(x, \phi) = \sup_{y \in \unitInt} |s(\mathscr{S}(x), y) - s(\phi, y)|.
\end{equation}
Since splines are continuous in their knot positions and heights, the function $E$ is also continuous in $\phi$ by the continuity of compositions of continuous functions.
Therefore, there exists a $\delta$ such that for all $x \in \unitInt^{d-1}$ and functions $\mathscr{S}'(x) \colon \unitInt^{d-1} \to \mathbb{R}^K$ such that $||\mathscr{S}'(x) - \mathscr{S}(x)||_{\infty} < \delta$ we have that
\begin{equation}
    E\left(x, \mathscr{S}'(x)\right) = \sup_{y \in \unitInt} |s\left(\mathscr{S}(x), y\right) - s\left(\mathscr{S}'(x), y\right)| < \epsilon / 3.
\end{equation}

We now use a classic universal approximation theorem for neural networks to show that a neural network exists that can be $\mathscr{S}'(x)$.
\citet{HORNIK1991251} showed that a single-hidden-layer feed-forward neural network can uniformly approximate any continuous function on a compact subset of $\mathbb{R}$, such as $\unitInt^{d-1}$.
Therefore, there exists feed-forward neural network $\mathscr{N} \colon \unitInt^{d-1} \to \mathbb{R}^K$ such that
\begin{equation}\label{eq:neural_net_approx}
    \sup_{x \in \unitInt^{d - 1}} ||\mathscr{N}(x) - \mathscr{S}(x)||_\infty < \delta.
\end{equation}
Finally, by combining Equations~\ref{eq:proof_spline_approx} and \ref{eq:neural_net_approx}, we have by the triangle inequality that 
\begin{align*}
    \sup_{(x, y) \in \unitInt^d}| s(\mathscr{N}(x), y) - f(x, y)| & \leq \sup_{(x, y) \in \unitInt^d} |f(x, y) - s(\mathscr{S}(x), y)| +
     \sup_{(x, y) \in \unitInt^d} |s(\mathscr{N}(x), y) - s(\mathscr{S}(x), y)|\\
    &  \leq 2 \epsilon / 3 < \epsilon.
\end{align*}
We have shown that a feed-forward network which outputs a spline can uniformly approximate any continuous function on the hypercube.
This completes the proof.

\end{proof}

\subsection{Proof of SPICE Interval Expressiveness}\label{app:intervals_proof}

\begin{theorem}\label{thm:app_intervals}
    Let $\mathscr{P} = \bigcup_{i=1}^m [a_i, b_i] \subseteq [0, 1]$ where $a_i < b_i < a_{i + 1}$ and $1 \leq m \leq M$ for some natural numbers $m$ and $M$.
    Then the following are true with $K = 4M + 2$:
    \begin{enumerate}
        \item Given any $\hat q \colon 0 < -\hat q \cdot |\mathscr{P}| < 1$ there exists a \methodnameNegDensity model with $K$ knots such that $\predSetFhat(x) = \mathscr{P}$ for some arbitrary $x$.
        \item Given any $0 < -\hat q < 1$ there exists a \methodnameHPD model with $K$ knots such that $\predSetHPD(x) = \mathscr{P}$ for some arbitrary $x$.
    \end{enumerate}
\end{theorem}

\begin{figure}[h]
\includegraphics[width=\linewidth]{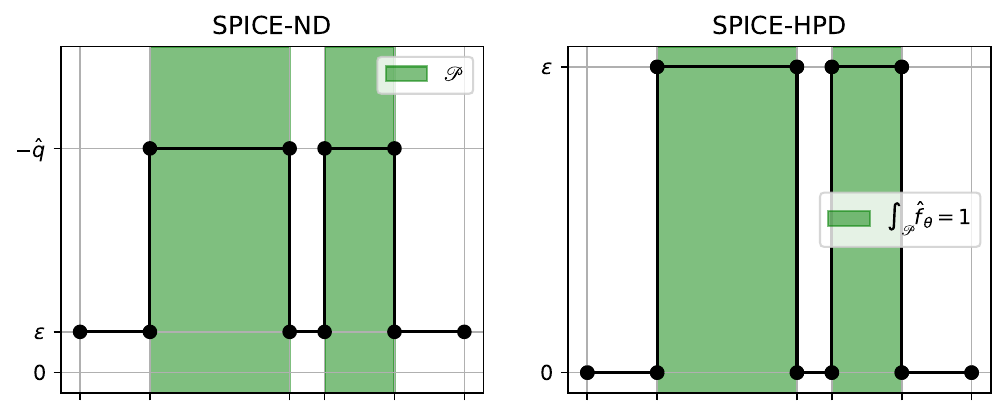}
\caption{Method for constructing an arbitrary union of intervals $\mathscr{P}$ as prediction sets by \methodnameNegDensity and \methodnameHPD.}
\label{fig:interval_proof}
\end{figure}

The proof of Theorem~\ref{thm:app_intervals} is a constructive proof and focuses on degree-one splines, since any higher degree spline can express a lower degree one.
The construction is not necessarily the most efficient way to build our target prediction set in terms of number of knots, but it shows that any such prediction set can be constructed with a linear number of knots.
See Figure~\ref{fig:interval_proof} for a visualization of the constructed prediction sets.

\begin{proof}
    We begin by noting that, as a result of Theorem~\ref{thm:app_universal}, \methodname can express any combination of knot positions $\knotpos$ and knot heights $\knotHeightLinear$ such that $\estimDensity$ integrates to one.
    Therefore, we will show that there exists a combination of knot parameters for any given $\hat q$ that results in the goal prediction set $\mathscr{P}$ and that results in $\estimDensity$ integrating to one.

    \paragraph{The \methodnameNegDensity case.}
    For \methodnameNegDensity, the prediction set is $\predSetFhat(x) = \{y \colon \estimDensity(y \mid x) > - \hat q \}$.
    We start by constructing the first interval $[a_1, b_1]$ using three knots.
    For convenience, we will define $c = - \hat q$.
    Set $(t_1, h_1) = (0, 0)$, $(t_2, h_2) = (a_1, 0)$, $(t_3, h_3) = (a_1, c + \epsilon)$, and $(t_4, h_4) = (b_1, c + \epsilon)$ for some $c > \epsilon > 0$.
    Note that the interval $\{\estimDensity(y \mid x) > c\ \colon y \in [0, b_1]\}$ is exactly $[a_1, b_1]$, and that $\int_0^{b_1} \estimDensity(y \mid x) dy = (c + \epsilon)(b_1 - a_1)$.
    This exact same procedure is then repeated for each interval, defining $\{t_i, h_i\}$ for $i = 1, \ldots, 4m$.
    Finally, we add $(t_{4m}, h_{4m})= (b_m, 0)$ and $(t_{4m + 1}, h_{4m + 1})= (b_m, 1)$.
    If we have more knots (i.e. $m < M$), we set them all to (1, 0) so they have no effect.
    We have now recreated $\mathcal{P}$.
    What remains is setting $\epsilon$ so that $\int \estimDensity(y \mid x)dy = 1$.
    Setting $\epsilon = \frac{1}{|\mathcal{P}|} - c$, the integral is:
    \begin{equation}
        \int_0^1 \estimDensity(y \mid x)dy = \sum_{i = 1}^m (b_i - a_i)(c + \epsilon) = (c + \epsilon) |\mathcal{P}| = 1,
    \end{equation}
    completing the \methodnameNegDensity case.

    \paragraph{The \methodnameHPD case.}
    \methodnameHPD requires new knot parameters.
    Set $(t_1, h_1) = (0, 0)$, $(t_2, h_2) = (0, a_i)$, $(t_3, h_3) = (a_1, \epsilon)$, and $(t_4, h_4) = (b_1, \epsilon)$ for some $0 < \epsilon$.
    Continuing this way, we set $(t_{4(i - 1) + 1}, h_{4(i - 1) + 1}) = (t_{4i}, 0)$, $(t_{4(i - 1) + 2}, h_{4(i - 1) + 2}) = (a_i, 0)$, $(t_{4(i - 1) + 3}, h_{4(i - 1) + 3}) = (a_i, \epsilon)$, and $(t_{4(i - 1) + 4}, h_{4(i - 1) + 4}) = (b_i, 0)$ for all $i = 1, \ldots, m$.
    If $m < M$, we set the remaining knots to $(1, 0)$ so they have no effect.
    Recall that the prediction set of \methodnameHPD is:
    \begin{equation}\label{app:hpd_predset}        
            \predSetHPD(x) = \left\{y \colon \int_{y' \colon \estimDensity(y' \mid x) \leq \estimDensity(y \mid x)} \estimDensity(y' \mid x)dy' > c \right\},
    \end{equation}
    where $c = - \hat q$.
    Since $\estimDensity = 0$ for any region outside of $\mathscr{P}$, we now no that $\predSetHPD(x)$ excludes all the regions out side of $\mathscr{P}$ as desired.
    Note that $\int \estimDensity(y \mid x) = \epsilon |\mathscr(P)|$.
    Setting $\epsilon = 1 / |\mathscr{P}|$ we satisfy the integration to one constraint.
    It is now also true that for any $y \in \mathscr{P}$, the integral $\int_{y' \in \mathscr{P})} \estimDensity(y' \mid x)dy' = 1$, since $\estimDensity(y \mid x) = \epsilon$ for all $y \in \mathscr{P}$.
    That means $\mathscr{P} = \predSetHPD(x)$ as desired.
    
\end{proof}

\section{NEURAL SPLINES}\label{app:cubic_spline}

\subsection{Knot Parameterization}\label{app:knot_encoding}
Throughout this section we will denote the encoding of $x$ as $\encoder(x) = z$.

\subsubsection{Knot Positions}
The knot position module maps $z$ to a sorted vector $\knotpos$ where $K$ is the number of knots.
We used a modified version of the code of \citet{durkan_cubic-spline_2019} to ensure the knot positions were sorted and between zero and one.
For numerical reasons, that involved using a minimum distance between knots, $\epsilon$.
The first step constructs unnormalized widths from $z$ using a single linear layer.
The unnormalized widths are then softmaxed to get widths that sum to one to get widths $v_i$.
Next, non-zero widths $w_i$ are constructed as:
\begin{equation}
    w_i = \epsilon + (1 - \epsilon * K)  v_i.
\end{equation}
Widths are converted to positions by taking the cumulative sum, and then prepending zero.
Finally, $\epsilon$ is added to the last knot position so that it equals one.

\methodname $(n = 2)$ requires one additional position parameter for each pair knot positions.
That position parameter is created by taking the midpoint between each consecutive pair of knot positions generated by the above procedure.

\subsubsection{\methodname $(n = 1)$ Knot Heights}
Unnormalized knot heights are produced using a single linear layer applied to $z$.
The heights are then made positive using the softplus function.
Finally, the heights are normalized by dividing the spline defined by the knot positions and heights by their integral (Appendix~\ref{app:integration}).

\subsubsection{\methodname $(n = 2)$ Knot Heights}
The knot heights for each non-midpoint position are the same as for $(n = 1)$, the softplus of a linear layer applied to $z$.
The knot heights for the midpoint positions are generated without using an activation, so they can be negative.
Lagrange coefficients ($a_i, b_i, c_i)$ which interpolate between each pair of consecutive positions and heights and the midpoint positions and heights, are computed.
These define the polynomial $a_i y^2 + b_i y + c$ on $[t_i, t_{i + 1}$.
After zero-truncation, these polynomials define a spline, which is then divided by its integral so it integrates to one.

\subsection{Spline Integration}\label{app:integration}

\subsubsection{Integrating \methodname $(n = 1)$}
Integrating a piecewise linear function just requires summing together the area of a series of trapezoids.
Each trapezoid's area is $\frac12 (t_{i + 1} - t_i) (h_i + h_{i + 1})$.

\subsubsection{Integrating \methodname $(n = 2)$}
Between each consecutive knot positions, we have a zero truncated polynomial $\bar p_i(y) = \max[a_i y^2 + b_i y + c, 0]$ on $[t_i, t_{i + 1}]$.
The first step is to integrate the non-zero-truncated polynomial $(a_i / 3 y^3 + b_i / 2 y^2 + c_i y) \biggr \rvert_{t_i}^{t_{i + 1}}$.
The next step is to compute the roots $r_i^1 < r_i^2$ (if they exist) using the quadratic formula.
If they exist, they are clipped to be in $[t_i, t_{i + 1}]$.
The same integral is then computed between the roots (which is necessarily negative) and then subtracted from the non-zero-truncated integral.

\subsection{Building Negative Density Prediction Sets}\label{app:nd_pred_set}

Given a conformal quantile $\hat q$, the first step is to compute all the intersection positions of each polynomial component $\estimDensity(y \mid x)$ and $- \hat q$.
The intersections are found by computing the roots of the polynomial $p_i(y) + \hat q$.
The roots are then clipped to be in $[t_i, t_{i + 1}]$.

In the $(n = 1)$ case, whether the interval is inside or outside the clipped roots is determined by the slope of polynomial.
In the $(n = 2)$ case, there are two possible cases to consider.

If the roots do not exist (are complex), then the interval is set to $[t_i, t_{i + 1}]$ if the vertex of the polynomial is above zero $(c_i - b_i^2 / (4  a) > 0)$.
Otherwise the interval is set to be empty.

If the roots do exist, the sign of the vertex is used to determine whether the prediction interval is between (positive vertex) or outside of the roots (negative vertex).
In the outside roots case, we get two intervals, which are $[t_i, r_i^1]$ and $[r_i^2, t_{i + 1}]$.

The intervals for each polynomial component are unioned to compute the prediction set $\predSetFhat(x)$.

\subsection{Building HPD Prediction Sets}\label{app:hpd_pred_set}
Given a conformal quantile $\hat q$, the prediction set is computed using the bisection algorithm.
A lower bound $\ell_0 = 0$ and $u_0 = \sup_y \estimDensity(y \mid x)$ are initialized.
Next a midpoint $m_i = (u_i + \ell_i) / 2$ is computed.
Next, the following integral is computed:
\begin{equation}
    F_i = \int_{\{y \colon \estimDensity(y \mid x) < m_i\}} \estimDensity(y \mid x) dy.
\end{equation}
If $F_i < - \hat q$, then $ell_{i + 1} = m_i$ and $u_{i + 1} = u_{i}$; otherwise $ell_{i + 1} = \ell_i$ and $u_{i + 1} = m_{i}$.
This process is repeated 15 times so that $F_15 \approx -\hat q$.
The HPD-prediction set is then $\{y \colon \estimDensity(y \mid x) > F_15\}$, which intervals can be computed the same way as for \methodnameNegDensity.

\section{EXPERIMENTAL DETAILS}

\subsection{Data Preprocessing}\label{app:preprocessing}
For each model, the training inputs ($x$) were preprocessed to have approximately zero mean and standard deviation one using the population mean and standard deviation estimated from the train split.

For \methodname models,  the targets, $y$, were processed to range approximately between zero and one using the scikit-Learn \citep{scikit} min-max scaler fit to the train split $y$s.

For the baseline models, besides Hist., the $y$s were transformed to have approximately zero mean and standard deviation one.

For Hist., $y$s were binned into discrete bins, with the number of bins being a hyperparameter.
The bin boundaries were evenly spaced between the minimum and maximum value of $y$.

\subsection{Shared Neural Network Encoder}\label{app:encoder}
Every model had a two-layer fully-connected neural network with GeLU \citep{gelu} activations to encode $x$.
The first layer mapped the input dimension to 32 hidden units.
The second layer mapped 32 hidden units to 32 encoding units.

Each model used depth one linear layers following the encoding to predict quantiles / bin probabilities / distribution parameters.

\subsection{Benchmark Datasets}\label{app:benchmark_datasets}
We give a brief description of each benchmark dataset and its source.
See \citet{romano_conformalized_2019} for more details about these data.

\begin{itemize}
    \item \textsc{bike}: UCI repository bike sharing data \citep{misc_bike_sharing_dataset_275} (CC BY 4.0 license).
    \item \textsc{BIO}: UCI repository \citep{misc_physicochemical_properties_of_protein_tertiary_structure_265} protein property prediction dataset (CC BY 4.0 license).
    \item \textsc{BLOG}: UCI repository \citep{misc_blogfeedback_304} blog feedback prediction dataset (CC BY 4.0 license).
    \item \textsc{MEPS19-21}: Medical expenditure panel survey datasets numbers 19 through 21 \citep{cohen2009MEPS} (Open Database License).
    \item \textsc{STAR}:  Student achievement dataset \citep{starDATA} (CC0 1.0 License).
    \item \textsc{TEMP.}: \citet{cho_temperature} introduced temperature forecasting data to compare machine learning models for correcting numerical weather simulations (CC BY 4.0 license).
\end{itemize}

\subsection{Data Splits}\label{app:data_split}
The data were randomly split into five buckets: train 50\%, validation 10\%, calibration 10\%, calibration validation 10\%, test 20\%.
The sample assignments within train / validation / calibration / validation calibration were determined by random seed, while the test set was held constant across all experiments and was only used to generate results after hyperoptimization.

\subsection{Hyperparameter Optimization}\label{app:hyperparameters}
Each model was hyperoptimized over a grid of five learning rates and four values of a capacity hyperparameter for a total of 20 combinations.
Each hyperparameter combination was evaluated on three different random seeds for robustness.

Evaluation was done by conformal calibration on the calibration split and conformal prediction on the calibration validation split.
Hyperparameters were selected uniquely for each dataset and model to minimize the average size at 90\% nominal coverage averaged across all three seeds.

The hyperparameter grids were as follows:
\begin{enumerate}
    \item \textbf{CQR}. Learning rate: $[1e-1, 5e-2, 1e-2, 5e-3, 1e-3]$. Nominal quantile interval: $ [0.3, 0.5, 0.7, 0.9]$.
    \item \textbf{CHR}. Learning rate: $[1e-1, 5e-2, 1e-2, 5e-3, 1e-3]$. Number of quantiles: $[50, 200, 350, 500]$.
    \item \textbf{PCP}. Learning rate: $[1e-1, 5e-2, 1e-2, 5e-3, 1e-3]$. Number of mixture components: $[5, 10, 15, 20]$.
    \item \textbf{Hist}. Learning rate: $[1e-1, 5e-2, 1e-2, 5e-3, 1e-3]$. Number of bins: $[11, 21, 31, 51]$.
    \item \textbf{SPICE $(n = 1)$}. Learning rate: $[5e-2, 1e-2, 5e-3, 1e-3, 5e-4]$. Number of knots: $[11, 21, 31, 51]$.
    \item \textbf{SPICE $(n = 2)$}. Learning rate: $[1e-2, 5e-3, 1e-3, 5e-4, 1e-4]$. Number of knots: $[11, 21, 31, 51]$.
\end{enumerate}

\subsection{Model Training}\label{app:training}
Models were trained using the AdamW \citep{adamw} optimizer with weight decay $1e-4$ for a maximum of 50,000 batches of size 512 stopping earlier if the validation loss did not improve after 125 passes over the training data.
The final model was selected based on its best validation loss, which was calculated every 100 batches or full pass over the training data, whichever was smaller. 
The validation loss was calculated on 10 batches or the full validation set, whichever was smaller.
The learning rate was decayed according to the cosine annealing schedule \citep{cosine_decay}.
The gradient was clipped to have max-norm of five.

\subsection{Conditional Histogram Baseline}\label{app:discretized_baseline}

The Hist. model was trained as a classifier model on discretized $y$ values (Appendix~\ref{app:preprocessing}) using the negative log-likelihood loss.
The architecture and hyperparameter optimization was shared with the other models.
The conformal score was the negative probability of the true class.

\subsection{Prediction Set Size Normalization}\label{app:size_normalization}
The size normalization can be thought of as dividing by the prediction set size of an unconditional version of the Hist. baseline.
$y$ was first discretized into 20 evenly spaced bins, then the probability of $y$ being in each bin was computed on the train split.
This resulted in an unconditional ``model'' for predicting $y \mid x$.
This model was then conformalized using the calibration split with the negative-probability of $y$ as the conformal score.
The score was then used to compute a constant prediction set for the test data.
The size of that prediction set was the normalization constant for that dataset.
All prediction set sizes in the results were divided by the corresponding normalization constant for the relevant dataset.
This process is in the function called \texttt{get\_baseline\_size} in the supplementary code.

\subsection{Label Conditional Coverage Approximation}\label{app:label_cond_coverage}
Label conditional coverage was approximated by binning $Y$ into five buckets.
Coverage was computed on the test set for each bucket.
Label conditional size was reported as the worst coverage across all five buckets.

\subsection{Synthetic Data Experiment}\label{app:synthetic_hetero}
In order to generate the synthetic bimodal data, 2,000 $x$ values were taken at even spacing between zero and one.
Noise was then sampled as
\[
	\sigma(x) = 0.1 + x U,
\]
where $U$ was random uniform on $[0, 1]$ for each $x$.
A random indicator was then sampled for each $x$ with equal probability as zero or one.
If the indicator was one, $y$ was set to $\sigma$.
Otherwise $y$ was set to $-\sigma$.
Preprocessing was done for each model the same way as for the benchmark data (Appendix~\ref{app:preprocessing}).

\end{document}